\documentclass{article}
\usepackage{spconf,amsmath,epsfig}
\usepackage{url}

\title{Remote Sensing for Weed Detection and Control}
\threeauthors 
    {Ishita Bansal}
    {Interlake High School\\ 16245 NE 24th St\\ Bellevue 98008, WA}
    {Peder Olsen}
    {Microsoft Research\\ 14820 NE 36th St\\ Redmond 98052, WA}
    {Roberto Estevão}
    {Microsoft\\ Rua Visconde de Inhauma, 83\\ Rio de Janeiro, Brazil}

\begin{document}
%
\maketitle
\begin{abstract}
Italian ryegrass is a grass weed commonly found in winter wheat fields that are competitive with winter wheat for moisture and nutrients.  Ryegrass can cause substantial reductions in yield and grain quality if not properly controlled with the use of herbicides.  To control the cost and environmental impact we detect weeds in drone and satellite imagery.  Satellite imagery is too coarse to be used for precision spraying, but can aid in planning drone flights and treatments. Drone images on the other hand have sufficiently good resolution for precision spraying.  However, ryegrass is hard to distinguish from the crop and annotation requires expert knowledge.  We used the Python \texttt{segmentation models} library \cite{Iakubovskii:2019} to test more than 600 different neural network architectures for weed segmentation in drone images and we map accuracy versus the cost of the model prediction for these.   Our best system applies herbicides to over 99\% of the weeds while only spraying an area 30\% larger than the annotated weed area.  These models yield large savings if the weed covers a small part of the field.  
\end{abstract}
\section{Introduction}
\label{sec:intro}
Several steps are involved in precision weed management. The farmer gathers surveillance imagery, manually annotates the weed locations, uploads the annotations to the spraying drone, and applies the herbicide.  Integrating a weed segmentation model into the sprayer drone allows all steps to happen with only one flight and without a time consuming manual annotation intervention.  Besides these benefits, this precision farming scenario can reduce the cost of herbicide, maintain the yield and reduce the environmental impact. 

When managing many fields, the time of flights as well as the weather must be taken into account.  It may not always be possible to treat all fields, thus the order of which fields are sprayed must be prioritized.   To assist with planning, satellite imagery could be used to measure the amount of ryegrass growing on each field.  As satellites have significantly lower resolution, we built a model that estimates the percentage of weed in each pixel.  This model still gives a good estimate of the total weed area, but the resulting weed annotation is inferior to that of a drone annotation and cannot directly inform the sprayer.

\begin{figure*}[htb]
\noindent
\begin{center}
\begin{minipage}[b]{0.24\linewidth}
\begin{center}
(a) UAV image\\
\includegraphics[width=0.9\linewidth]{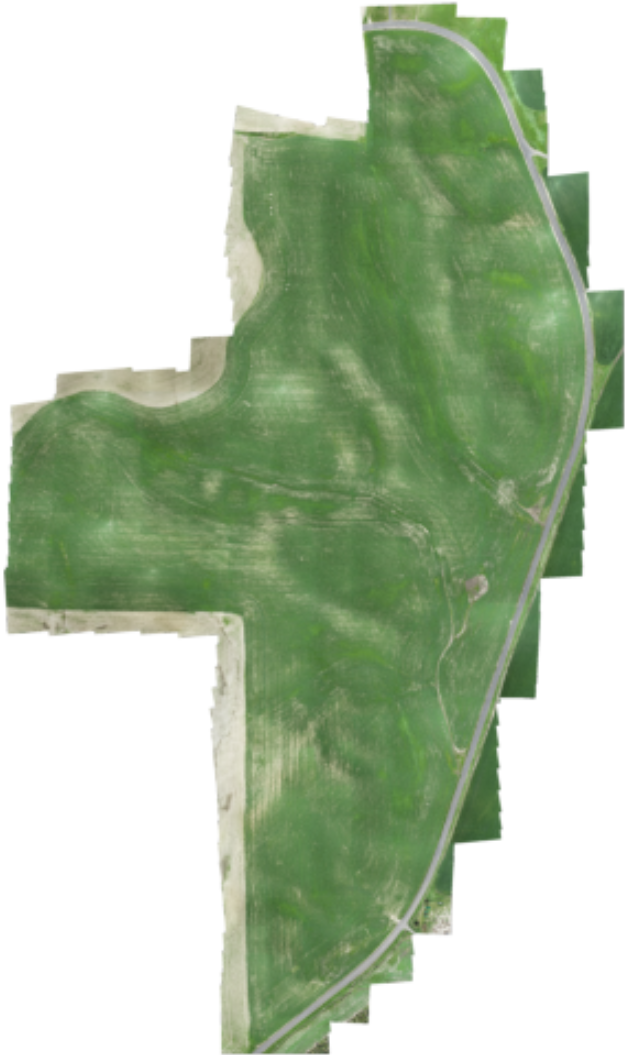}
\end{center}
\end{minipage}%
\begin{minipage}[b]{0.24\linewidth}
\begin{center}
(b) Weed annotation\\
\includegraphics[width=0.9\linewidth]{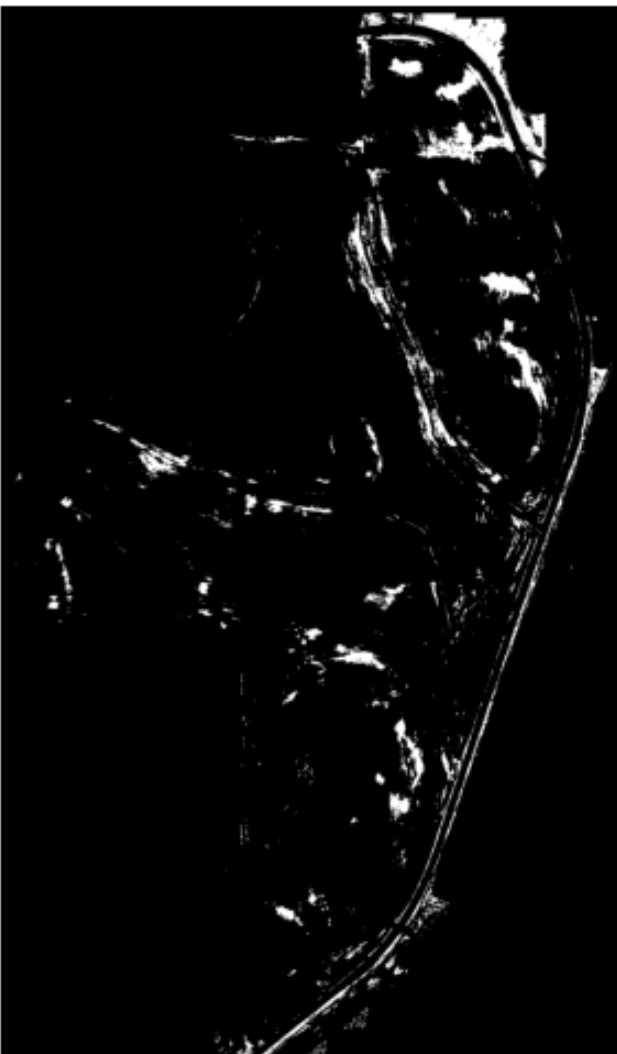}
\end{center}
\end{minipage}%
\begin{minipage}[b]{0.24\linewidth}
\begin{center}
(c) Sentinel-2 RGB image\\
\includegraphics[width=0.9\linewidth]{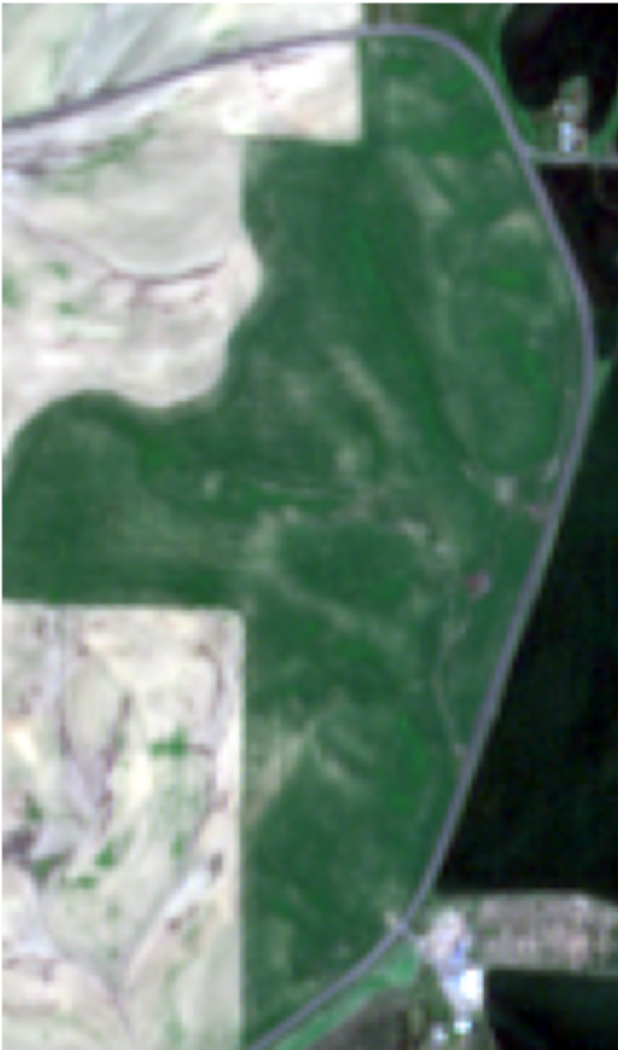}
\end{center}
\end{minipage}%
\begin{minipage}[b]{0.24\linewidth}
\begin{center}
(d) False color image\\
\includegraphics[width=0.9\linewidth]{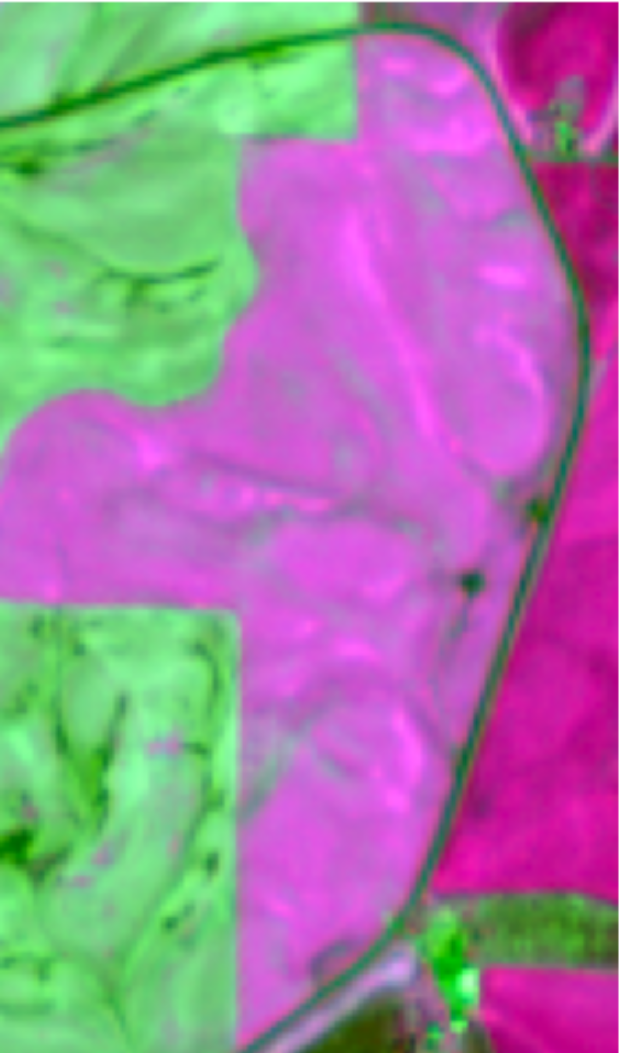}
\end{center}
\end{minipage}
\end{center}
\caption{A drone image (a), the corresponding weed annotation (b), a Sentinel-2 image of the location (c) and a false color composite using NIR, green and Vegetation Red Edge Band 2 in place of the RGB bands.}\label{fig:field}
\end{figure*}

Figure \ref{fig:field} shows a field of winter wheat from the perspective of a drone and a satellite.  Comparing the the mask in Figure~\ref{fig:field}~(b), we can see that the ryegrass has a slightly lighter green color than the crop in Figure~\ref{fig:field}~(a) and (c).  However, the Sentinel-2 satellite provides other spectral bands as well among which are 3 vegetation red edge bands and a near infrared band.  We used the second vegetation red edge band, the green band and the near infrared band to make the false color composite in Figure~\ref{fig:field}~(d).  In this image the ryegrass is showing up in bright pink and the contrast between the ryegrass and the winter wheat is much greater than in the RGB image.  In what follows we describe our data set and show experimental for both the satellite and drone weed detection models.

\section{Data Description}
We had approximately $1.3\times 10^{9}$ annotated pixels to use for training, held out and test data.  45\% of the data was assigned to training, 25\% as held out data and 30\% as test data.  We used the held out data to tune hyperparameters and evaluated models on the test data.  For the satellite image annotation, we changed the projection and aligned the drone image to match the Sentinel-2 image.  Sentinel-2 images have a 10m$\times$10m pixel resolution while the drone image had a pixel resolution of 2.9cm$\times$2.9cm.  We resampled the drone image to 5cm$\times$5cm resolution in Sentinel-2's projection.  Each Sentinel-2 pixel thus corresponds to 200$\times$200 drone pixels from which we calculated the percentage of drone pixels that were annotated as weeds. The percentage pixel values formed a soft mask that we use as a regression target.  The total area of weed cover can be computed as the pixel area times the percentage.  The Sentinel-2 products were corrected for bottom of the atmosphere (L2A products) and there were 4 bands at 10m$\times$10m resolution (blue, green, red and near infrared (NIR)), 6 bands at 20m$\times$20m resolution (3 vegetation red edge bands (VRE), 1 narrow near infrared (NNIR), 2 short wave infrared bands (SWIR)), and 2 atmospheric observation bands at 60m$\times$60m resolution.  We did not utilize the lowest resolution bands and the 20m$\times$20m resolution bands were resampled to 10m$\times$10m resolution to ease processing.

\section{Satellite Pixel Models}
Each pixel is a 10-dimensional vector from which we are trying to predict the percentage of area covered in weed.  This is a regression problem and there are regressors that we can train for this problem.  Scikit-learn, \cite{scikit-learn}, provides a multitude of such models from which we tuned 42. The 10 models with the highest Pearson correlation coefficient ($R^2$) are listed in Table~\ref{tab:r2}.  The best performing model had an $R^2$ of 0.589.  From these 10 models, we considered all subsets of 3 models for use in the ensemble model \texttt{VotingRegressor} with uniform weights. The best model combination used \texttt{NuSVR}, \texttt{SVR} and \texttt{ExtraTreesRegressor} with an $R^2$ of 0.641. We then optimized the weights yielding an $R^2$ of 0.643, which is a minute improvement on the held-out data. Table~\ref{tab:test} shows the performance of these 3 top models on the test data, and we see that the small improvement for the weighted model does not hold on the test data.  Nevertheless, the difference is small and we proceed with the weighted ensemble model.

\begin{table}[htb]
\small
\begin{tabular}{lccc}
{\bf Model} & {\bf RMSE} & {\bf MAE} & $R^2$ \\
\hline
\texttt{NuSVR} &                0.0879 &  0.0436 & {\bf 0.5886}  \\ 
\texttt{SVR} &                  0.0911 &  0.0497 & 0.5841  \\
\texttt{MLPRegressor} &         0.0918 &  0.0501 & 0.5605  \\
\texttt{BaggingRegressor} &     0.0924 &  0.0565 & 0.5581  \\
\texttt{KNeighborsRegressor} &  0.0947 &  0.0400 & 0.5510  \\
\texttt{ExtraTreesRegressor} &  0.0951 &  0.0450 & 0.5440  \\
\texttt{HistGradientBoostingReg.} & 0.0981 &  0.0336 & 0.4350  \\
\texttt{RandomForestRegressor} & 0.0995 &  0.0438 & 0.4350  \\
\texttt{TheilSenRegressor} &    0.1073 &  0.0460 & 0.4120  \\
\texttt{CCA} &                  0.1132 &  0.0842 & 0.4034  \\
\end{tabular}
\caption{The top 10 best regressors from the perspective of the Pearson correlation coefficient.}\label{tab:r2}
\end{table}

\begin{table}[htb]
\small
\begin{tabular}{lccc}
{\bf Model} & {\bf RMSE} & {\bf MAE} & $R^2$ \\
\hline
\texttt{NuSVR} & 0.1116 &  0.0596 & 0.5416  \\ 
\texttt{VotingRegressor} & 0.1093 & 0.0590 & {\bf 0.5955}\\
weighted \texttt{VotingRegressor} & 0.1085 & 0.0584 & 0.5858\\
\end{tabular}
\caption{Comparison of models on the test data.}\label{tab:test}
\end{table}

The test data covered a total of 50 acres of land. The weed covered 2.45 acres (4.85\%), while the model predicts a total of 2.83 weed acres.  For the held out data the weed covered 2.90 acres, with the model predicting 2.97 acres.  Another way to evaluate the performance of the model is to see how much land would have to be sprayed to guarantee a large fraction of the weed is covered in pesticide.  We pick a threshold for the minimum predicted weed-fraction and spray all Sentinel-2 pixels with a larger predicted weed-fraction.  Table~\ref{tab:spraying-frac} shows the performance when targeting 90\%, 95\% and 98\% of the weeds.  We see that from satellite images we end up greatly over spraying even when the actual amount of weeds is relatively small.  This is not due to the model quality, but due to having to spray an entire pixel even if it contains as little as 10\% weeds.  The more spread out the weeds the poorer the performance of the spraying.  We will see that with the higher resolution of drone images this is no longer the case in the next section.

\begin{table}[htb]
\small
\begin{center}
\begin{tabular}{lcccc}
{\bf Threshold} & {\bf Weed \%} & {\bf Land \%} & {\bf Land Acres} & {\bf Excess \%}\\
\hline
0.036 & 90\% & 42\% & 21 & 766\%\\ 
0.025 & 95\% & 53\% & 27& 984\%\\
0.012 & 98\% & 68\% & 34& 1300\%\\
0.001 & 99\% & 83\% & 41& 1590\%\\
\end{tabular}
\end{center}
\caption{Amount of acres sprayed as a function of the predicted weed-fraction threshold. The columns are respectively the threshold used, the actual percentage of weeds sprayed, the percentage of land sprayed, the total acres sprayed and the excess percentage sprayed compared to an oracle model.}\label{tab:spraying-frac}
\end{table}

\section{Drone Segmentation Models}
The Python library \texttt{segmentation models}~\cite{Iakubovskii:2019} provides a framework for building a multitude of segmentation models for practically any application.  It provides 9 different model architectures, each allowing for a choice of more than 624 encoders that can be used with many of the model architectures, and 8 loss functions that can be used for the training.  This allows training more than 5000 distinct models with 7 loss functions (one loss function is for multi-class segmentation).  We afforded ourselves the luxury of training $\approx$ 500 different combinations of losses and models and we sampled all loss functions as well as all the available 124 encoders (and none of the additional $\geq 500$ \texttt{timm} \cite{rw2019timm} encoders).   With the U-net architecture~\cite{ronneberger2015u} as the main model architecture, we found that, among the 7 loss functions, the \texttt{Focal} \cite{lin2017focal}, BCE (\texttt{BCEWithLogits}) and the SoftBCE (\texttt{SoftBCEWithLogits}) \cite{Khvedchenya_Eugene_2019_PyTorch_Toolbelt} loss functions worked well, as can be seen from Table~\ref{tab:losses}.  The four remaining loss functions lagged significantly behind when required to spray 99\% of the weeds, but performed adequately in the 90\% weed coverage category.  The \texttt{Focal} loss performed best for 90, 95 and 98\% weed coverage, while \texttt{BCEWithLogits} loss performed best for 99\% weed coverage.   

\begin{table}[htb]
\small
\begin{center}
\setlength{\tabcolsep}{4pt}
\begin{tabular}{lccccc}
{\bf Loss} & {\bf Encoder} & \multicolumn{4}{c}{\bf Excess}\\
 &  & 90\% & 95\%  & 98\% & 99\% \\
\hline
BCE & \texttt{VGG19} &  -3.8\% & 6.7\% & 19.3\% & {\bf 29.2\%} \\
\texttt{Focal} & \texttt{VGG19} &  {\bf -4.6\%} & {\bf 6.0\%} & {\bf 19.3\%} & 29.8\% \\
SoftBCE & \texttt{VGG16} &  -4.2\% & 6.1\% & 20.0\% & 31.5\% \\
\texttt{Lovasz} & \texttt{VGG16} &  -3.3\% & 6.3\% & 22.0\% & 51.3\% \\
\texttt{Tversky} & \texttt{DenseNet169} &  0.3\% & 12.1\% & 71.4\% & 413\% \\
\texttt{Dice} & \texttt{DenseNet169} &  -2.6\% & 10.0\% & 91\% & 435\% \\
\texttt{Jaccard} & {\scriptsize\texttt{TIMM\_REGNETX\_002}} &  -0.24\% & 12.3\% & 98\% & 1195\% \\
\end{tabular}
\end{center}
\caption{Best trained model using the UNET model architecture for each loss function.  The columns shows the excess spraying area measured as a percentage of the total weed area for spraying respectively 90\%, 95\%, 98\% and 99\% of the weed.}\label{tab:losses}
\end{table}

Table~\ref{tab:performance} shows a select few of the models that were trained.  The best model used UNET++ as a model architecture~\cite{zhou2018unet++} and merely over-sprayed by 28\% for a 99\% weed coverage. The FPN model architecture \cite{lin2017feature} also performs well, while some of the other larger models that are designed for more complex segmentation tasks (such as people and faces) are less effective.  This is logical as Italian ryegrass is recognized very much from the color and texture, hence local context suffices for detection.

It should be noticed that the VGG class of models (VGG16 originated from \cite{simonyan2014very}), fairs quite well. Another notable model listed uses UNET with the \texttt{TIMM\_REGNETX\_002} encoder \cite{radosavovic2020designing}. It runs more than 8 times faster than the UNET++ model with the \texttt{VGG19} encoder and the size of the model is 19MB as compared to 179MB for the best model.  From a practical point of view this may be a very good model choice as the spraying drone is not capable of centimeter or pixel level precision when applying the herbicide.  The increase in the spraying area compared to the best model is after all merely 0.25 acres out of a total of 50 field acres, and a smaller faster model is more suitable for edge computations on the sprayer.

\begin{table*}[htb]
\small
\begin{center}
\begin{tabular}{lllllcccc}
{\bf Model} & {\bf Encoder} & {\bf Loss} & {\bf Size} & {\bf Speed} & \multicolumn{4}{c}{\bf Excess}\\
 & & & & & {\bf 90\%} & {\bf 95\%}  & {\bf 98\%} & {\bf 99\%} \\
\hline
UNET++ & \texttt{VGG19} & BCE & 179MB & 1.0 & -4.11\%  & 6.30\%  & 18.63\%  & {\bf 28.09\%} \\
UNET++ & \texttt{VGG16} & \texttt{Focal} & 158MB & 1.3 & -3.74\%  & 6.98\%  & 18.99\% & 28.36\% \\
UNET++ & \texttt{VGG16} & SoftBCE & 158MB & 1.5 & -4.24\%  & 6.14\%  & {\bf 18.43\%}  & 28.45\% \\
UNET & \texttt{VGG19} & BCE & 116MB & 4.4 & -3.77\%  & 6.73\%  & 19.33\%  & 29.22\% \\
UNET & \texttt{VGG19} & \texttt{Focal} & 116MB & 3.0 & {\bf -4.63\%}  & {\bf 6.04\%}  & 19.30\%  & 29.82\% \\
FPN & \texttt{VGG16} & \texttt{Focal} & 67MB & 3.0 & -3.10\%  & 8.35\%  & 21.59\%  & 31.83\% \\
UNET & \texttt{MIT\_b0} & SoftBCE & 22MB & 1.6 & -2.58\%  & 8.45\%  & 21.98\%  & 33.02\% \\
UNET & \texttt{VGG13} & BCE & 74MB & 5.06 & -1.95\%  & 9.73\%  & 23.48\%  & 34.63\% \\
FPN & \texttt{MIT\_b0} & \texttt{Focal} & 20MB & 2.55 & -2.65\%  & 9.25\%  & 23.72\%  & 35.11\% \\
FPN & \texttt{MIT\_b0} & BCE & 20MB & 2.58 & -1.54\%  & 11.00\%  & 26.14\%  & 37.90\% \\
UNET & \texttt{VGG11} & BCE & 73MB & 6.18 & -1.68\%  & 11.03\%  & 26.51\%  & 38.84\% \\
UNET & \texttt{TIMM\_REGNETX\_002} & BCE & {\bf 19MB} & {\bf 8.46} & -1.56\%  & 10.83\%  & 26.58\%  & 39.85\% \\
\end{tabular}
\end{center}
\caption{A select list of trained models.  The speed is measured relative to the top model.}\label{tab:performance}
\end{table*}

\begin{figure}[htb]
\begin{center}
\includegraphics[width=0.95\linewidth]{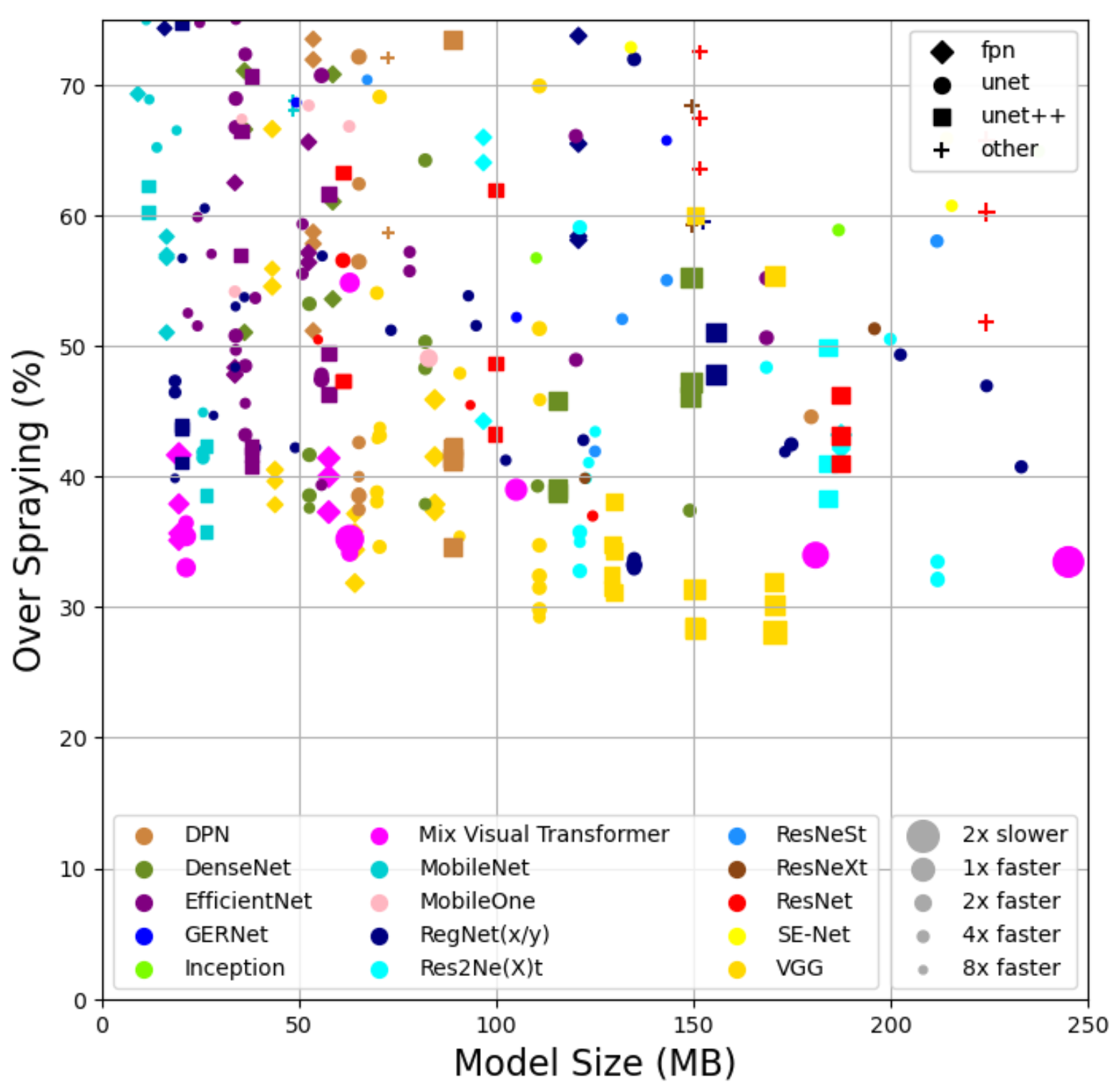}
\end{center}
\caption{Performance of many of the trained models.  The performance is shown on the y-axis, the model-size on the x-axis, while the size of the marker indicates the inference time, the color indicates the encoder group and the marker shape the model architecture.}\label{fig:performance}
\end{figure}

Figure~\ref{fig:performance} presents a plot of the models where the size of the markers indicates the evaluation time of the model while the color indicates the encoder group (e.g. \texttt{VGG} for \texttt{VGG11}, \texttt{VGG13}, \texttt{VGG16}, etc.) of the model.  While only select models are detailed in Table~\ref{tab:performance}, Figure~\ref{fig:performance} provides a wider view of the performance landscape for those wanting to train their own weed detection models.

\begin{figure}[htb]
\setlength{\tabcolsep}{1pt}
\begin{tabular}{cccc}
{\bf Zoom out} & {\bf Segmentation} & {\bf Annotation} & {\bf Prediction}\\ 
\includegraphics[width=0.23\linewidth]{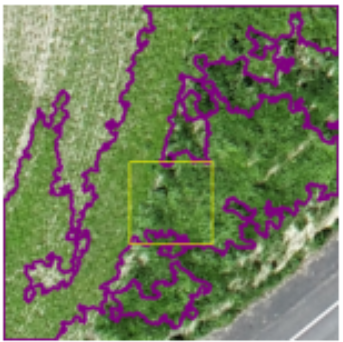} &
\includegraphics[width=0.23\linewidth]{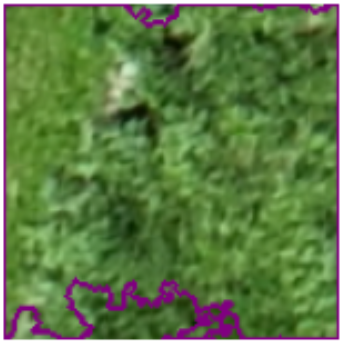} &
\includegraphics[width=0.23\linewidth]{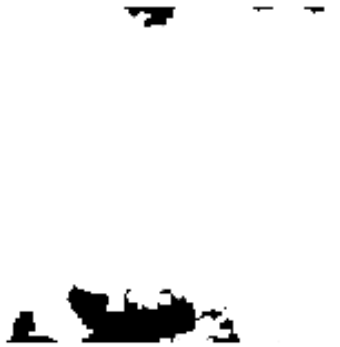} &
\includegraphics[width=0.24\linewidth]{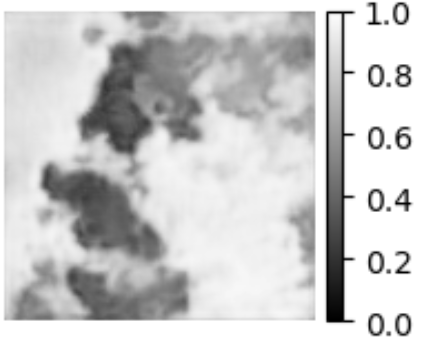} \\
\includegraphics[width=0.23\linewidth]{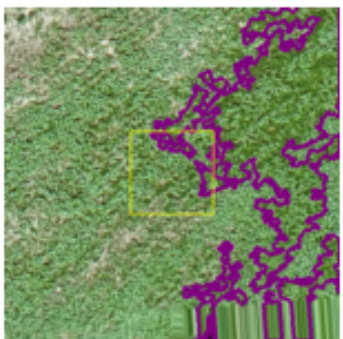} &
\includegraphics[width=0.23\linewidth]{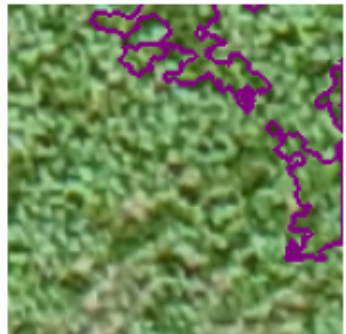} &
\includegraphics[width=0.23\linewidth]{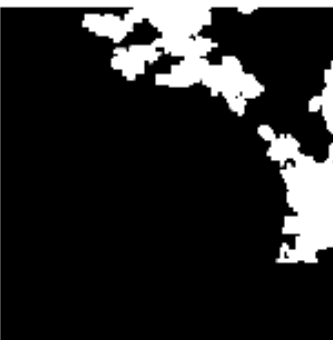} &
\includegraphics[width=0.24\linewidth]{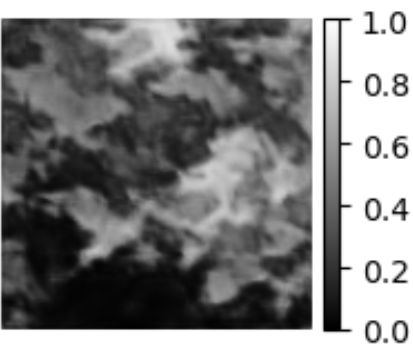} \\
\includegraphics[width=0.23\linewidth]{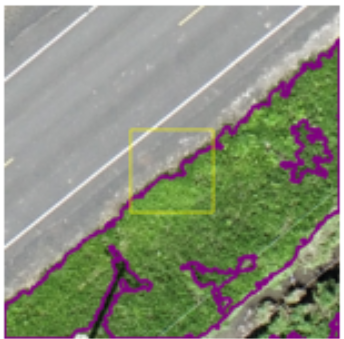} &
\includegraphics[width=0.23\linewidth]{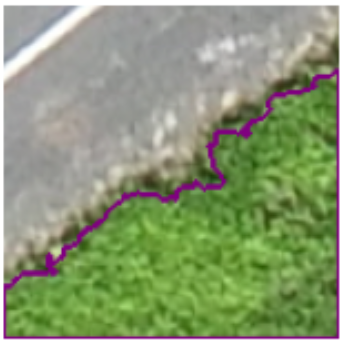} &
\includegraphics[width=0.23\linewidth]{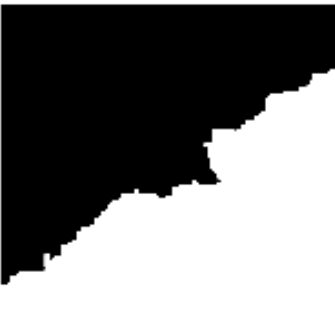} &
\includegraphics[width=0.24\linewidth]{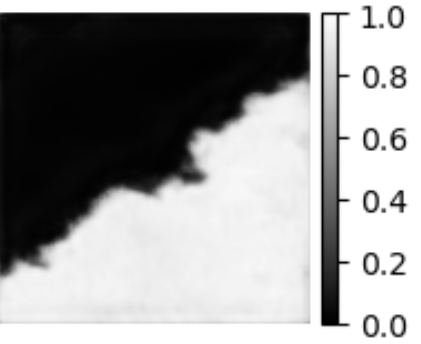} \\
\includegraphics[width=0.23\linewidth]{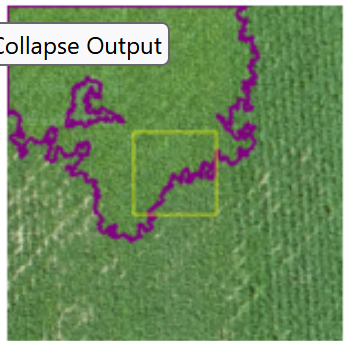} &
\includegraphics[width=0.23\linewidth]{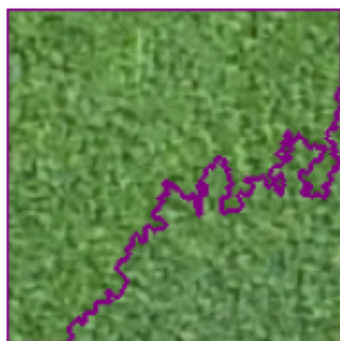} &
\includegraphics[width=0.23\linewidth]{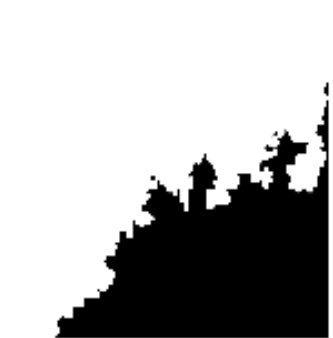} &
\includegraphics[width=0.24\linewidth]{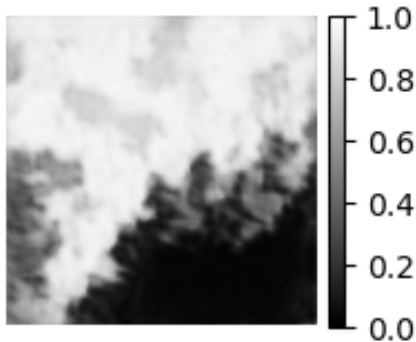} \\
\end{tabular}
\caption{Examples of weed segmentation in drone imagery. Left to right: input with segmentation boundaries (zoom out), input with segmentation boundaries, segmentation mask, and model prediction.}\label{fig:examples}
\end{figure}

Finally, Figure~\ref{fig:examples} offers some example visualizations for the weed detection.  We show (left to right): a zoomed out view of the input with the overlaid model prediction, a closer look at the input and segmentation boundaries, the segmentation mask, and the raw prediction as a float. The segmentation masks shown used the minimal threshold needed to spray 99\% of the weeds on the held-out data. We can see from the last column that the model is not as confident in cases where the weed doesn't stand out very well (e.g. the second row).

\section{Conclusions}
In this work, we presented our weed detection model that accurately and efficiently detects ryegrass present in farmer's crop fields.  We showed how to use segmentation models for precision weed control and satellite imagery for weed treatment planning.  
In future research we will expand the data sets to help increase the accuracy. Larger data sets can help optimize this model, to expand to a variety of weed species and to be geographically more diverse. We also explored using satellite imagery to spectrally extend the drone imagery to bands that gives stronger contrast between the weed and the crop.  Initial experiments did not improve our results.  A possible reason being that the ground truth annotation is inaccurate and biased by the RGB imagery.  This can be fixed by annotating using multispectral drone images.

\clearpage

\bibliographystyle{IEEEbib}
\bibliography{refs}

\end{document}